\documentclass{article}

\PassOptionsToPackage{numbers, compress}{natbib}

\usepackage{PRIMEarxiv}
\usepackage{graphicx}
\usepackage{subcaption}

\usepackage[utf8]{inputenc} 
\usepackage[T1]{fontenc}    
\usepackage{hyperref}       
\usepackage{url}            
\usepackage{booktabs}       
\usepackage{amsfonts}       
\usepackage{nicefrac}       
\usepackage{microtype}      
\usepackage{xcolor}         
\usepackage{xpatch}
\usepackage{natbib}

\usepackage{xpatch}
\makeatletter
\xapptocmd{\NAT@bibsetnum}{\setlength{\leftmargin}{0pt}\setlength{\itemindent}{\labelwidth}\addtolength{\itemindent}{\labelsep}}{}{}
\makeatother

\title{CIPHER: Scalable Time Series Analysis for Physical Sciences with Application to Solar Wind Phenomena}

\author{%
 Jasmine R. Kobayashi\thanks{These authors contributed equally to this work.}\\
 Southwest Research Institute \\
 \texttt{jasmine.kobayashi@swri.org} \\
 \And
 Daniela Martin\footnotemark[1] \\
 University of Delaware \\
 \texttt{dmartinv@udel.edu}
 \And
 Valmir P. Moraes Filho \\
 Catholic University of America \\
 \texttt{moraesfilho@cua.edu} \\
 \And
 Connor O'Brien \\
 Boston University \\
 \texttt{obrienco@bu.edu} \\
 \And
 Jinsu Hong \\
 Georgia State University \\
 \texttt{jhong36@gsu.edu} \\
 \And
 Sudeshna Boro Saikia \\
 Universität Vienna \\
 \texttt{sudeshna.boro.saikia@univie.ac.at} \\
 \And
 Hala Lamdouar \\
 University of Oxford \\
 \texttt{lamdouar@robots.ox.ac.uk} \\
 \And
 Nathan Miles \\
 University of Colorado Boulder \\
 \texttt{nathan.miles-1@colorado.edu}
 \And
 Marcella Scoczynski \\
 Universidade Tecnológica Federal do Paraná \\
 \texttt{marcella@utfpr.edu.br} \\
 \And
 Mavis Stone \\
 Stanford University \\
 \texttt{moraesfilho@cua.edu} \\
 \And
 Sairam Sundaresan \\
 Intel Labs \\
 \texttt{sairam.sundaresan@intel.com} \\
 \And
 Anna Jungbluth \\
 European Space Agency \\
 \texttt{anna.jungbluth@t-online.de} \\
 \And
 Andrés Muñoz-Jaramillo \\
 Southwest Research Institute \\
 \texttt{amunozj@boulder.swri.edu} \\
 \And
 Evangelia Samara \\
 NASA Goddard Space Flight Center \\
 \texttt{evangelia.samara@nasa.gov} \\
 \And
 Joseph Gallego \\
 Drexel University \\
 \texttt{jg3959@drexel.edu} \\
}

\begin{document}

\maketitle

\begin{abstract}
Labeling or classifying time series is a persistent challenge in the physical sciences, where expert annotations are scarce, costly, and often inconsistent. Yet robust labeling is essential to enable machine learning models for understanding, prediction, and forecasting. We present the \textit{Clustering and Indexation Pipeline with Human Evaluation for Recognition} (CIPHER), a framework designed to accelerate large-scale labeling of complex time series in physics. CIPHER integrates \textit{indexable Symbolic Aggregate approXimation} (iSAX) for interpretable compression and indexing, density-based clustering (HDBSCAN) to group recurring phenomena, and a human-in-the-loop step for efficient expert validation. Representative samples are labeled by domain scientists, and these annotations are propagated across clusters to yield systematic, scalable classifications. We evaluate CIPHER on the task of classifying solar wind phenomena in OMNI data, a central challenge in space weather research, showing that the framework recovers meaningful phenomena such as coronal mass ejections and stream interaction regions. Beyond this case study, CIPHER highlights a general strategy for combining symbolic representations, unsupervised learning, and expert knowledge to address label scarcity in time series across the physical sciences. The code and configuration files used in this study are publicly available to support reproducibility.
\end{abstract}

\section{Introduction} 
Time series data is central to many domains in science and engineering, from finance to space sciences \cite{fu2011review, almeida2023time}. Labeling parts of the time series is a costly process that usually involves a human expert in the loop \cite{Ratner_2017,Sun_2017}. In the physical sciences, such labels are critical for identifying key phenomena that support theory validation, model training, and space weather forecasting. Yet generating consistent annotations across massive datasets remains prohibitively expensive \cite{Song_2022}. Beyond this scalability issue, a central challenge is the scarcity of expert annotations in physics domains: only a handful of specialists are typically available to provide labels, and their judgments can vary, leading to inconsistencies across datasets. This scarcity makes it difficult to build reliable labeled corpora at the scale required by modern machine learning approaches. Nowadays, the growth of large-scale sensing networks and continuous monitoring platforms has led to an unprecedented expansion in the volume and resolution of available data \cite{aggarwal2015data}. While this wealth of data enables new discoveries, it also poses significant challenges for storage, analysis, and interpretation, particularly for clustering and pattern recognition tasks. Traditional clustering algorithms often struggle with scalability when applied to massive collections, as the number of comparisons increases linearly, as in \textit{k}-means, or linearithmically, as in DBSCAN \cite{ran2023comprehensive}. Compression techniques provide a practical solution by transforming raw time series into lower-dimensional, symbolic representations, preserving key dynamical patterns while reducing computational complexity \cite{camerra-etal-2010}. When combined with density-based clustering methods, such as Hierarchical DBSCAN (HDBSCAN) \cite{campello-etal-2013}, these representations enable the discovery of recurring phenomena across large datasets that would otherwise be intractable.  

Solar wind time series illustrate this challenge. From the long-running OMNI dataset \cite{king-papitashvili-2005, omniweb} to high-resolution measurements by NASA’s Parker Solar Probe \cite{raouafi2023parker}, decades of continuous observations exceed hundreds of gigabytes. Analyzing these data is critical for understanding solar wind phenomena and space weather \cite{temmer2021space}. In this context, any framework capable of grouping and labeling data should not only handle scale, but also maximize the utility of limited expert time by focusing labeling efforts on representative subsets.

In this work, we introduce the \textit{Clustering and Indexation Pipeline with Human Evaluation for Recognition} (CIPHER), which integrates \textit{indexable Symbolic Aggregate approXimation} (iSAX) compression \cite{camerra-etal-2010}, HDBSCAN clustering \cite{campello-etal-2013}, a human-in-the-loop labeling/classification processing, and a final propagation of the set of human labels to the entire clusters. This method can label/classify large-scale physics datasets. Applied to solar wind phenomena, CIPHER can recover well-known phenomena such as coronal mass ejections (CMEs) and stream interaction regions (SIRs), underscoring its effectiveness in identifying meaningful events within large, noisy time series. More broadly, CIPHER provides a general strategy for combining symbolic compression, unsupervised clustering, and expert-in-the-loop validation to overcome label scarcity in physics domains. While we demonstrate its utility in heliophysics, the same principles can be applied to other areas of the physical sciences, such as seismology, plasma physics, or climate research, where complex time series are abundant but expert annotations remain sparse.

\section{Model: CIPHER}
CIPHER consists of four main steps. The first step is preprocessing, which includes an optional detrending and smoothing feature to remove large-scale biases and high-frequency noise \cite{monke2020optimal}, as well as normalization of the time series. Subsequences are then compressed using iSAX \cite{camerra-etal-2010}, which transforms Piecewise Aggregate Approximation coefficients into symbolic words via statistically defined breakpoints. Each time series is segmented into fixed-length windows, a parameter referred to as the “chunk size,” and further subdivided into smaller segments, called the “word size,” which determine the temporal resolution of the symbolic encoding. This multi-resolution representation enables scalable indexing while preserving essential temporal patterns.

Clustering is performed on selected levels of the iSAX index using HDBSCAN \cite{campello-etal-2013}, a density-based algorithm that groups similar sequences while labeling low-density regions as noise. One hyperparameter determines the minimum number of sequences that must be grouped together for the algorithm to consider them a valid cluster; this parameter is called “min\_cluster\_size.” Another hyperparameter controls how strictly the algorithm treats points in low-density regions as noise, referred to as “min\_samples.” To handle unassigned points, the pipeline can optionally re-cluster them under relaxed density constraints. The final step involves a domain expert evaluating representative windows from each cluster to validate physical consistency, resolve ambiguities, and assign meaningful labels. This human-in-the-loop step allows propagating expert annotations across the entire cluster.

\section{Experimental Setup}
\begin{figure}[htbp]
\centering
\includegraphics[width=1.\linewidth]{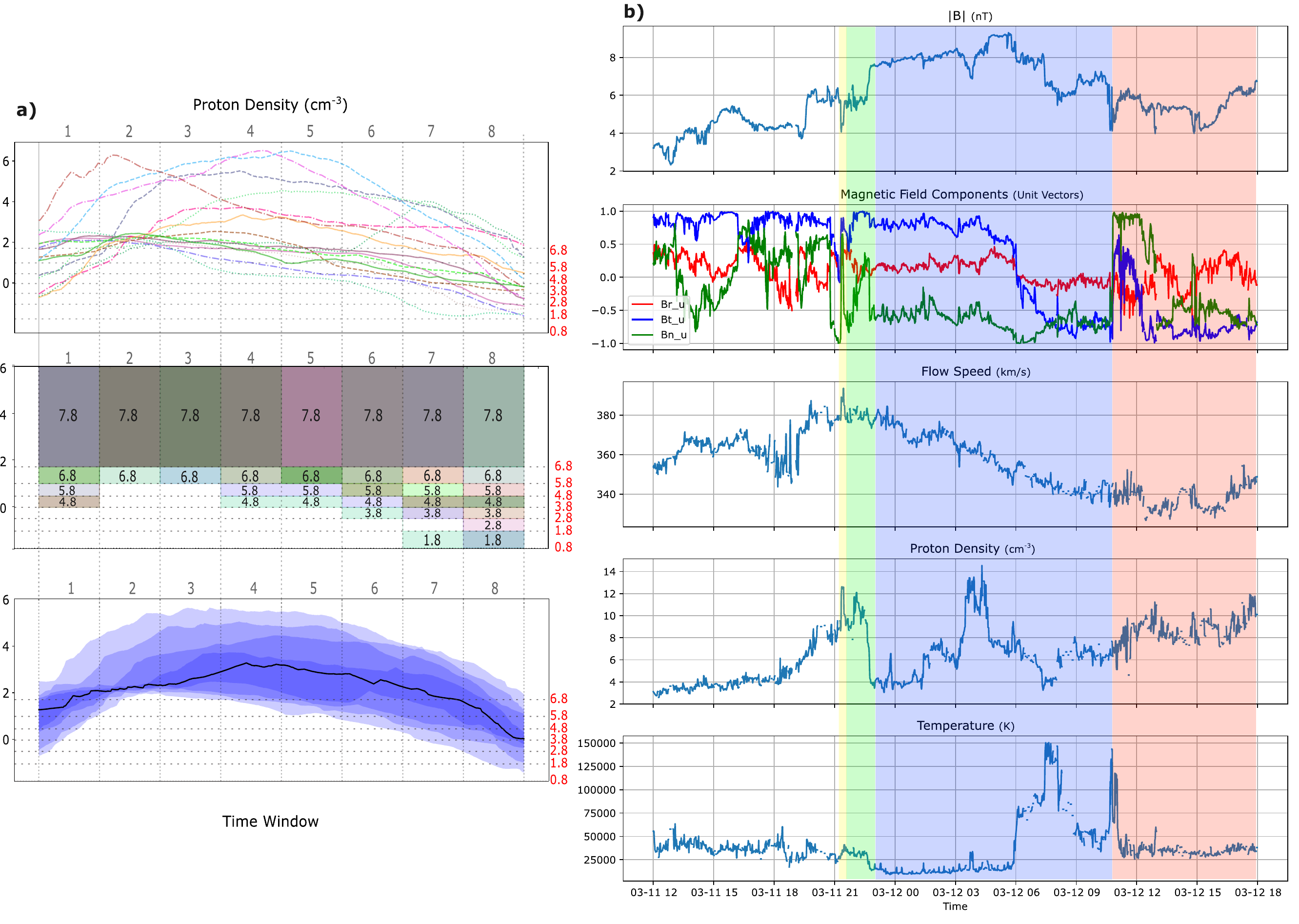}
\caption{\textbf{\textit{a)}} Structure of a CIPHER cluster built from smoothed and detrended proton density sequences with a 35-hour window. \textbf{Top:} individual preprocessed time series assigned to the cluster. \textbf{Middle:} symbolic representations obtained via iSAX compression, showing the index values that defined the cluster membership. \textbf{Bottom:} summary statistics of the cluster, with shaded confidence intervals (5–95\%) in purple and the black curve representing the mean sequence.
\textbf{\textit{b)}} Raw solar wind data of one representative sequence from the cluster (Subfigure \textbf{a}), capturing the March 11–12, 2021 CME \citep{Möstl_2020, helioforecast_icmecat}. The five panels display different OMNI parameters (total magnetic field, magnetic field components, flow speed, proton density, and temperature) observed simultaneously. These complementary signatures were jointly analyzed by the domain expert to confirm the CME classification. Shaded regions highlight the key substructures that guided this expert validation: \textbf{yellow}: forward shock, \textbf{green}: compressed sheath, \textbf{blue}: magnetic ejecta, and \textbf{red}: the trailing solar wind.}
\label{fig:experiment-fig1}
\end{figure}
We evaluate CIPHER on solar wind data, a key observable for heliophysics and space weather research. This section outlines the dataset, method configuration, and training procedure.

\textbf{OMNI Data.} The OMNI dataset provides a long-term record of solar wind conditions compiled from multiple spacecraft (ACE, WIND/SWE, IMP-8, Geotail) \citep{king-papitashvili-2005}. Measurements are propagated to the Earth’s first Lagrange point (L1) \cite{eldo2024review}, corresponding to the upstream solar wind at the bow shock. Its standardized frame and extensive temporal coverage make OMNI a cornerstone for space weather studies. In this work, we use 1-minute resolution OMNI data, focusing on bulk flow speed, proton density, proton temperature, and magnetic field components.

\textbf{Configuration and Training Procedure.} For the experiments reported in this paper, we used a chunk size of 35 hours and a word size of 8 for the iSAX compression. HDBSCAN clustering was configured with a minimum cluster size of 5 and a noise-sensitivity parameter also set to 5. These values were selected as they provided a balance between capturing meaningful phenomena and maintaining computational efficiency. Additional parameters and sensitivity tests are reported in the Supplemental Material.

\section{Results and Discussion}
We found that CIPHER is capable of clustering events; in particular, when applied to solar wind data, it successfully grouped distinct solar phenomena such as CMEs and SIRs. Figure \ref{fig:experiment-fig1} (a) depicts a cluster identified by CIPHER using smoothed and detrended proton density sequences. We focused on proton density because it exhibits strong and recurrent signatures during CME passages, making it a robust parameter for automated clustering. To validate this cluster, a domain expert examined a subset of sequences using the raw solar wind data. Crucially, the expert did not rely on density alone but cross-checked additional parameters, ensuring that the cluster assignment was physically consistent. Figure \ref{fig:experiment-fig1} (b) shows one representative sequence from this subset, capturing the March 11–12, 2021 CME \citep{Möstl_2020, helioforecast_icmecat}. The analysis highlights well-defined CME substructures \citep{Richardson_2010,Kilpua_2017,Kilpua_2015,Jian_2006}, including the forward shock, compressed sheath, magnetic ejecta, and trailing solar wind; demonstrating that CIPHER can leverage simple clustering inputs while still enabling systematic expert validation across multiple physical dimensions. Similar validation workflows were carried out for SIR-related clusters, following the same procedure of combining clustering outputs with expert inspection to ensure the physical consistency of the assignments.

\begin{figure}[htbp]
\centering
\includegraphics[width=1.\linewidth]{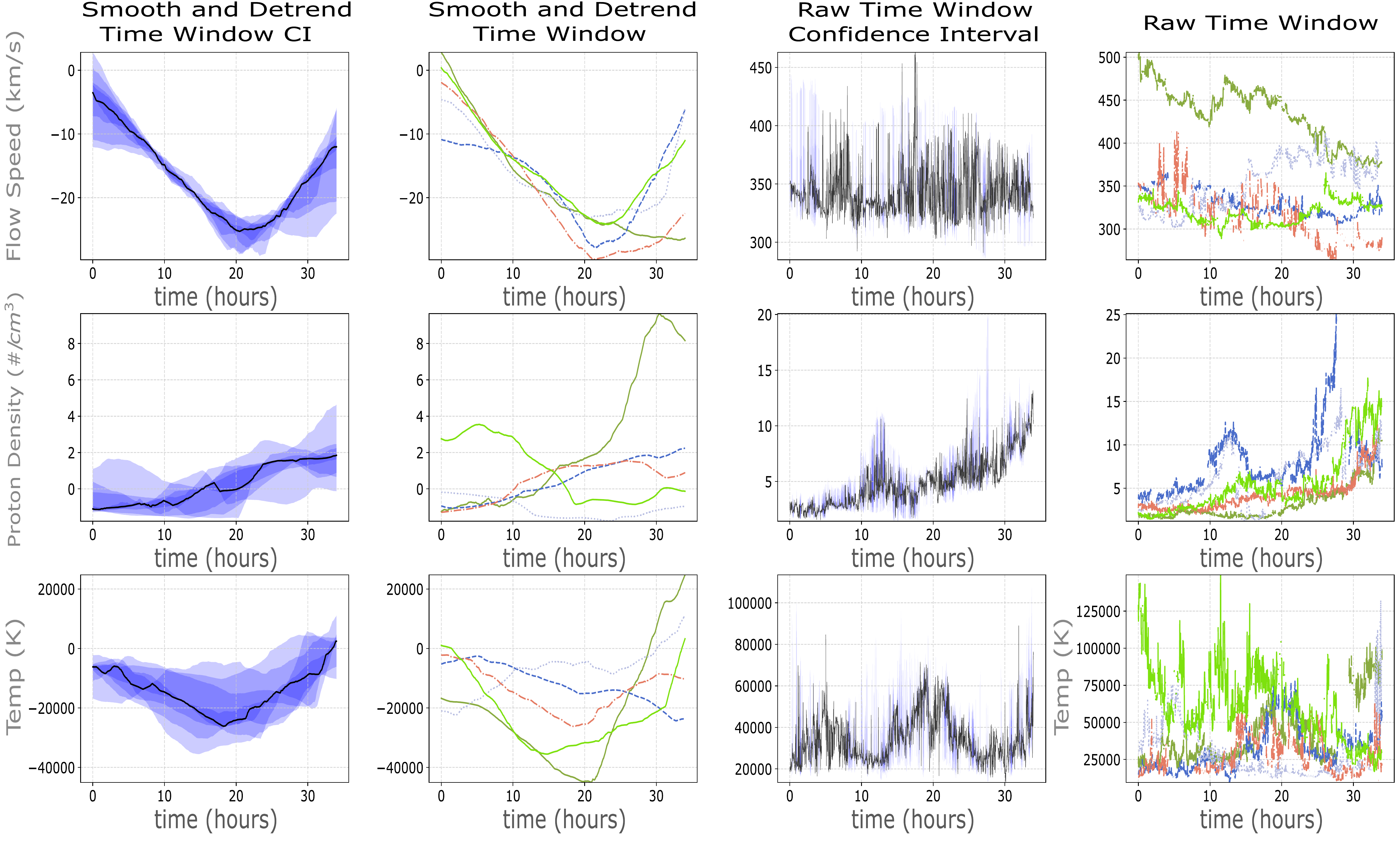}
\caption{CIPHER cluster associated with a stream interaction region (SIR), discovered from solar wind flow speed data.
\textbf{Rows:} Flow speed (top), proton density (middle), and temperature (bottom).
\textbf{Columns:} From left to right: (1) confidence intervals (5–95\%) of preprocessed data (smoothed and detrended, with mean in black); (2) individual preprocessed sequences; (3) confidence intervals of the corresponding raw sequences(5–95\%; with mean in black); (4) individual raw sequences.
The comparison highlights the contrast between the noisy raw measurements and the structured coherence revealed by preprocessing. While raw time series appear too irregular for manual grouping, the narrow confidence intervals of the preprocessed data show that these parameters evolve consistently across events, confirming the cluster’s physical meaning as validated by the domain expert.}
\label{fig:experiment-sir-multiparam}
\end{figure}

Figure \ref{fig:experiment-sir-multiparam} shows that, despite the high variability in the raw data, the preprocessed sequences reveal clear and coherent patterns across all three parameters. This setup was designed as a targeted experiment to demonstrate that CIPHER can identify coherent clusters based on one primary parameter (flow speed) while cross-validating the corresponding time intervals in additional parameters (proton density and temperature), even when the raw data appear too noisy to group manually. The narrow confidence intervals in the preprocessed views indicate that the time series genuinely belong to the same physical category. By contrast, the raw sequences alone appear too irregular for a human observer to recognize such grouping, underscoring the ability of CIPHER to extract meaningful patterns that would be otherwise hidden in the data. 

\section{Conclusion}
CIPHER provides a scalable, interpretable framework for labeling complex time series in the physical sciences, combining symbolic compression, density-based clustering, and expert-in-the-loop validation. Applied to solar wind data, the method successfully recovered meaningful phenomena, including CMEs and SIRs, demonstrating that representative expert annotations can be efficiently propagated across large datasets. These results illustrate that CIPHER can reveal coherent structures in noisy, high-dimensional data, enabling systematic and reproducible classification while reducing dependence on exhaustive manual labeling.

\textbf{Limitations and Future Work.} While CIPHER accelerates expert-driven labeling, its performance depends on the choice of primary clustering parameters and the availability of domain experts for initial validation. Current experiments focused on a limited set of solar wind parameters; extending the framework to incorporate joint multi-parameter clustering or additional modalities could improve sensitivity to subtle phenomena. Future work will explore automated selection of compression and clustering hyperparameters, integration with streaming data, and applications to other domains of the physical sciences with scarce expert annotations.

\section*{Broader Impact}
CIPHER addresses a central bottleneck in scientific research: the scarcity and cost of expert-labeled time series. By combining symbolic representations, scalable clustering, and human-in-the-loop validation, the framework enables rapid, systematic classification of complex physical phenomena. In heliophysics, this allows efficient identification of CMEs and SIRs, supporting space weather forecasting and operational decision-making. Beyond solar wind studies, CIPHER offers a generalizable approach for any domain with abundant, complex time series and limited expert availability, including seismology, plasma physics, and climate science. By reducing reliance on exhaustive manual labeling, CIPHER can accelerate scientific discovery while maintaining interpretability and physical rigor. The full codebase supporting this work is available at \url{https://github.com/spaceml-org/CIPHER}.

\section*{Acknowledgements}
This work is a research product of Heliolab (heliolab.ai), an initiative of the Frontier Development Lab (FDL.ai). FDL is a public–private partnership between NASA, Trillium Technologies (trillium.tech), and commercial AI partners including Google Cloud and NVIDIA. Heliolab was designed, delivered, and managed by Trillium Technologies Inc., a research and development company focused on intelligent systems and collaborative communities for Heliophysics, planetary stewardship and space exploration. We gratefully acknowledge Google Cloud for extensive computational resources enabled through VMware. This material is based upon work supported by NASA under award No. 80GSFC23CA040. Any opinions, findings, and conclusions or recommendations expressed are those of the author(s) and do not necessarily reflect the views of the National Aeronautics and Space Administration.

\small
\bibliographystyle{unsrt}  
\bibliography{references}

\normalsize

\end{document}